%% file: main.tex
\documentclass[10pt,twocolumn,letterpaper]{article}

\usepackage{cvpr}              %

\input{preamble}

\definecolor{cvprblue}{rgb}{0.21,0.49,0.74}
\usepackage[pagebackref,breaklinks,colorlinks,allcolors=cvprblue]{hyperref}
\usepackage{amssymb}
\usepackage{array,multirow}

\def\paperID{12} %
\def\confName{CVPR}
\def\confYear{2025}

\title{On the Importance of Conditioning for Privacy-Preserving Data Augmentation}

\author{Julian Lorenz \quad Katja Ludwig \quad Valentin Haug \quad Rainer Lienhart\\
University of Augsburg\\
Augsburg, Germany\\
{\tt\small \{julian.lorenz,katja.ludwig,valentin.haug,rainer.lienhart\}@uni-a.de}
}

\begin{document}
\maketitle
\input{sec/0_abstract}    
\input{sec/1_intro}

\input{sec/2_related_work}

\input{sec/3_method}

\input{sec/4_results}

\input{sec/5_blackbox}

\input{sec/6_conclusion}    
{
    \small
    \bibliographystyle{ieeenat_fullname}
    \bibliography{main}
}

\end{document}

%% file: preamble.tex
\newcommand{\todo}[1]{{\color{red}#1}}

%% file: sec/0_abstract.tex
\begin{abstract}
Latent diffusion models can be used as a powerful augmentation method to artificially extend datasets for enhanced training.
To the human eye, these augmented images look very different to the originals.
Previous work has suggested to use this data augmentation technique for data anonymization.
However, we show that latent diffusion models that are conditioned on features like depth maps or edges to guide the diffusion process are not suitable as a privacy preserving method.
We use a contrastive learning approach to train a model that can correctly identify people out of a pool of candidates.
Moreover, we demonstrate that anonymization using conditioned diffusion models is susceptible to black box attacks.
We attribute the success of the described methods to the conditioning of the latent diffusion model in the anonymization process. The diffusion model is instructed to produce similar edges for the anonymized images. Hence, a model can learn to recognize these patterns for identification.
\end{abstract}

%% file: sec/1_intro.tex
\section{Introduction}
\label{sec:intro}

In recent years, neural networks have grown significantly in size. However, training large models effectively requires extensive datasets to prevent overfitting. While advancements in unlabeled pretraining have reduced the need for massive datasets, a substantial amount of data is still necessary. This poses a challenge for machine learning tasks involving humans, such as human pose estimation or face recognition, where collecting large-scale datasets is difficult. Obtaining comprehensive consent from all individuals in custom datasets is often impractical, further complicating data acquisition.
\begin{figure}[t]
  \centering
   \includegraphics[width=\linewidth]{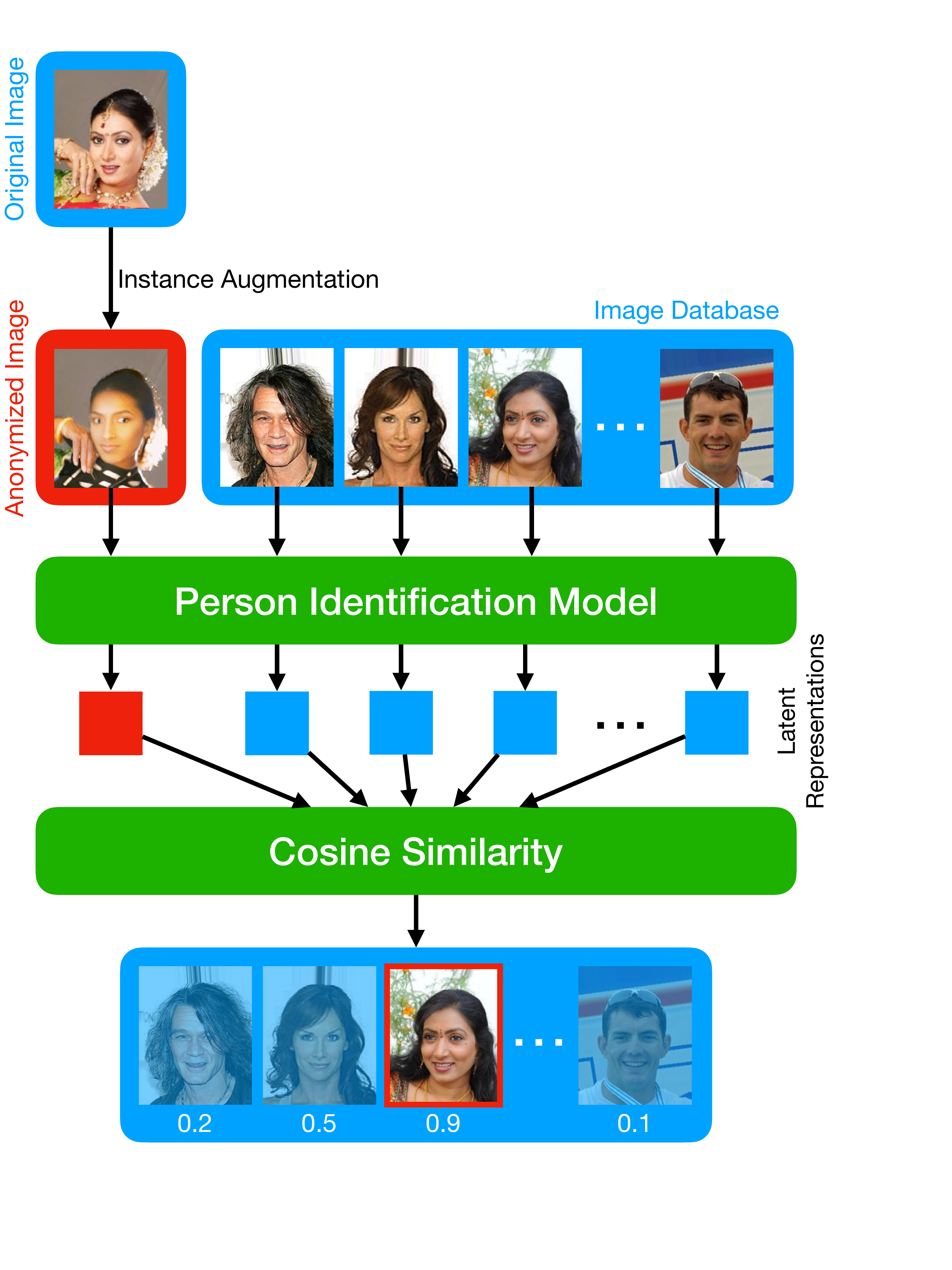}

   \caption{Overview of our method. An original image is anonymized with Instance Augmentation. Our model takes this image and an image database as an input and outputs latent representations for all images, encoding the person identity. We calculate the cosine similarity scores for the anonymized representation vector and the representations of all persons in the image database and select the image with the highest similarity.}
   \label{fig:method}
\end{figure}

A potential solution to this challenge is privacy-preserving data augmentation, which aims to obscure individuals' identities by replacing the parts of images showing real humans with synthetically generated images while keeping the background. This approach preserves the diversity of real-world (in-the-wild) datasets while ensuring anonymity. However, these models are not entirely foolproof and may still leak sensitive information, and in the worst case, an attacker can reveal the identity of the original person.

In this paper, we examine the risk of information leakage in privacy-preserving data augmentation methods that utilize conditioned latent diffusion models.
Recent work \cite{instance_aug} introduces an augmentation technique called \emph{instance augmentation} that uses pre-trained latent diffusion models to redraw sensitive parts of an image.
The authors claim that augmented faces can only be identified \cite{arcface} in 0.14\% of all cases for the COCO dataset \cite{coco}.
Although this is a great achievement and indeed enhances privacy, it is not sufficient to fully preserve privacy as we demonstrate in this work.

We evaluate the following scenario:
Given an anonymized image generated by an instance-level augmentation procedure (e.g., \cite{instance_aug}), an attacker attempts to determine whether it contains a specific individual. In this scenario, the attacker possesses a set of non-anonymized reference images, which also contain images of the target person. Successfully revealing the individual’s identity can have serious consequences. For instance, if the person appears in sensitive datasets - such as those related to protests, medical research, or controversial contexts —exposure could lead to harassment, blackmail, or doxxing. In fields like healthcare or banking, unauthorized access to private medical or financial data could result in fraud or discrimination.

In this paper, we demonstrate that even a simple neural network architecture is sufficient to identify individuals in a dataset anonymized using instance-level augmentations. To achieve this, we leverage SimCLR \cite{simclr}, a contrastive learning method, to train a model that generates latent representations encoding a person's identity from the input image. The model is trained to produce similar latent representations for images of the same individual while ensuring distinct representations for different individuals.
To identify a person in the dataset, we compute cosine similarity between the latent vectors of the anonymized input image and those of the attacker’s image database. Using this approach, we successfully re-identify 69.7\% of individuals in our test dataset, which contains 3,418 different people.

We hypothesize that the success of our method stems from the preservation of edges and depth in the instance augmentation process. To test this assumption, we attempt to reveal individuals’ identities using only edge images. This scenario can also be viewed as a black-box attack, where the attacker has no access to the augmentation pipeline or any augmented images.
To train our model, we generate edge images using either a HED detector \cite{hed} or a Canny edge detector \cite{canny}, without applying any anonymization. Notably, this approach does not require the attacker to have knowledge of the augmentation pipeline. After training, we apply the model to identify individuals in anonymized images. First, the anonymized images are converted into edge representations using the same detector as in training. Then, we compute similarity scores as in the standard setting. A visualization of our method can be found in \cref{fig:method}.
On a dataset containing 3,418 individuals, our black-box method successfully identifies the correct person in an anonymized image with a probability of 25.1\%.

To further investigate our hypothesis, we conduct an ablation study on the conditioning of the instance augmentation method, which can preserve edges, depth, or only the segmentation mask. Our experiments show that preserving either edges or depth alone is sufficient for our attack to succeed. Notably, depth preservation inherently retains edge information, as edges are a key component of facial depth. A similarity analysis of edge images across different augmentation variants further supports this hypothesis.

All these findings suggest that even simple methods can effectively re-identify individuals in instance-level augmented datasets if edge information is retained. This raises significant concerns about the use of such techniques for privacy-preserving anonymization.

\begin{figure*}[t]
  \centering
  \includegraphics[width=0.8\linewidth]{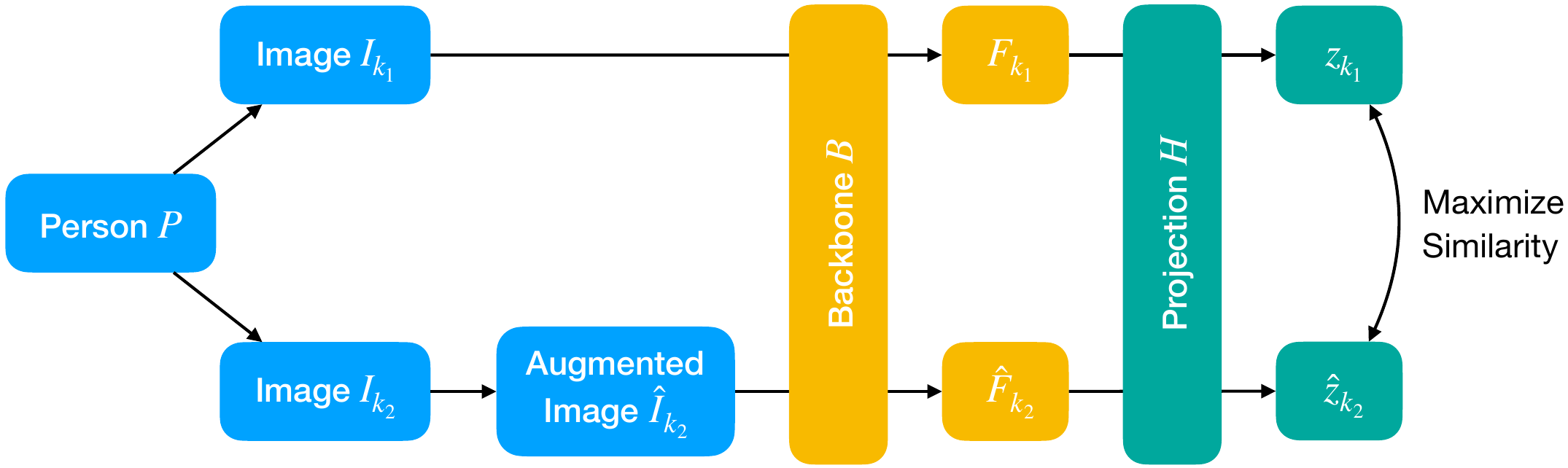}
   \caption{Our contrastive learning framework. An image $I_{k_1}$ and an augmented image $\hat{I}_{k_2}$ originating from the same person are passed through the backbone $B$ and the projection head $H$. The similarity of the resulting representations $z_{k_1}$ and $\hat{z}_{k_2}$ is maximized during training.}
   \label{fig:simclr}
\end{figure*}

However, instance-level augmentation remains a powerful technique for improving model performance. It effectively expands datasets with more diverse images while preserving corresponding annotations.
Nonetheless, as our work demonstrates, such methods are not suitable for anonymization. The primary issue lies in the conditioning of the latent diffusion model. Since the model is conditioned to reconstruct the original edges and/or depth map, a neural network can learn this correspondence. This learned information can then be exploited for privacy-revealing attacks.

Our contributions can be summarized as follows:

\begin{enumerate}
\item We identify an information leak in data anonymization frameworks that rely on instance-level augmentation techniques using conditioned latent diffusion models.
\item We demonstrate that contrastive learning is a simple yet highly effective method for identifying individuals in an anonymized dataset.
\item We hypothesize that the preservation of edges is the primary factor driving the success of our method and provide supporting evidence through ablation studies, where we train our model on differently augmented images.
\item We extend our approach by introducing a black-box method that does not require access to the anonymization pipeline yet still successfully identifies individuals.
\end{enumerate}

We will publish our code, weights, and the generated datasets upon paper acceptance.

%% file: sec/2_related_work.tex
\section{Related Work}
\label{sec:related_work}

\subsection{Privacy-Preserving Methods and Attacks}

Protecting the privacy for sensitive content in images has been a longstanding challenge in computer vision, with methods including pixelation, blurring \cite{privacy_blurring} or other classical approaches like P3 \cite{privacy_p3}.
However, methods that fundamentally alter the underlying image have a usability tradeoff \cite{viewer_experience} as parts of the image might not be recognizable anymore.
Generative models provide a solution to this problem. Instead of merely destroying information, they can replace parts of the image with generated data \cite{singh_tooner,privacy_imggen,privacy_a3gan,privacy_gan,9828699,hukkelås2019deepprivacygenerativeadversarialnetwork,face_deidentify}.

However, improvements on neural network architectures also facilitate attack methods \cite{anonymization_survey}.
Prior work has developed methods to defeat classical methods \cite{img_attack1}, but also more sophisticated methods have been rendered ineffective by deep learning methods \cite{img_attack2,medical_attack,attack_survey,face_attack}. %

\subsection{Instance Augmentation}

In this work we re-evaluate the anonymization capabilities of \textit{Dataset Enhancement with Instance-Level Augmentations} \cite{instance_aug}.
In their work, the authors present a data augmentation method that uses latent diffusion models to redraw selected instances in an image.
To ensure instance replacements that still fit the associated ground truth, they condition the diffusion process using ControlNet \cite{controlnet} and T2I Adapter \cite{t2i_adapter}.
These methods ensure that the segmentation mask, depth map, and edge map of the augmented images are similar to their original counterparts.
They show that their augmentation strategy improves performance on object detection, semantic segmentation, and salient object detection.

Additionally, the augmentation method is evaluated as a data anonymization tool.
To this end, ArcFace \cite{arcface} is employed to identify peoples' faces in the COCO dataset \cite{coco}. On this evaluation, only 0.14\% of all faces can be identified.
However, we argue that the reported numbers are misleading because evaluating anonymization requires a different benchmark.
Instead, a face dataset (\eg \cite{merler_faces,msceleb,celeba}) with multiple images of the same person provides a much more realistic evaluation benchmark as we demonstrate in this work.

\subsection{Contrastive Learning}

Contrastive Learning is a form of self-supervised learning \cite{9462394}. %
 The basic idea is to learn pairwise similarity of images and dissimilarity to other images contained in the batch, depending on the architecture, the extracted features of the convolutional neural network can then be compared with another feature output under a similarity metric, to decide how similar they are deemed by the neural net \cite{technologies9010002}. Recent implementations of this principle are MoCo \cite{He_2020_CVPR}, SwAV \cite{NEURIPS2020_70feb62b} and SimCLR \cite{chen2020simpleframeworkcontrastivelearning}, with a version of the last one being utilized for our research.

%% file: sec/3_method.tex
\section{Method}

In this section, we describe the simple contrastive learning framework that we use as our method, which is based on SimCLR \cite{simclr}. We further describe the three different evaluation protocols that we use, which cover different attacker strengths. 

\subsection{Contrastive Learning}

Our method is a contrastive learning approach similar to SimCLR \cite{simclr}. During training, we use pairs of original images $I$ and corresponding augmented images $\hat{I}$ that originate from the same person $P$. Note that $\hat{I}$ was created using an original image of the same person $P$, but definitely a different image. We feed a batch of $K$ images $I_k, k=1,...,k$ and a batch of images $\hat{I}_k$ through our backbone $B$, resulting in features $F_k$ and $\hat{F}_k$. Each image $I_k$ in the batch shows a different person. The features $F_k$ and $\hat{F}_k$ are then projected with a small projection head $H$ onto a lower-dimensional representation $z_k$ and $\hat{z}_k$. The network is now trained to maximize the similarities between the matching representations $z_k$ and $\hat{z}_k$ while pushing apart the similarities of the representations $z_k$ and $\hat{z}_j, k\neq j$ of different persons. SimCLR \cite{simclr} has shown that such a projection performs better than computing the similarity of the features $F_k$ and $\hat{F}_k$. 

Formally, let the similarity function $s(x, y)$ be defined as
\begin{equation}
    s(x, y) = \frac{x^Ty}{\lVert x\rVert \lVert y\rVert}.
\end{equation}
The loss function for a pair of matching representations is then defined as

\begin{equation}
\begin{split}
    &L(z_k, \hat{z}_k) = \\&- \log \frac{\exp(s(z_k, \hat{z}_k)/\tau}{\sum_{j=1}^{K} \textbf{1}_{k \neq j} \exp(s(z_k, z_j)/\tau) + \exp(s(z_k, \hat{z}_j)/\tau)},
\end{split}
\end{equation}
with $\textbf{1}_{k \neq j}$ evaluating to 1 if $k\neq j$ and to 0 otherwise. $\tau$ is a temperature parameter, which either pushes the similarities further apart or pulls them together depending on its value. Figure \cref{fig:simclr} visualizes the contrastive learning framework. The usage of a different base image for the augmented version is shown by using the image indices $k_1$ and $k_2$ in the figure.

\subsection{Evaluation}\label{sec:eval}

To evaluate our methods, we use three different evaluation protocols which we call the \emph{full reference evaluation}, the \emph{single reference evaluation}, and the the \emph{few references evaluation}. We will explain all protocols in the following.

\paragraph{Full Reference Evaluation (Full-Ref).} In this evaluation scenario, we suppose that the attacker has a database with multiple images of the same person. The similarity of the augmented image's representation from our model to the representations from all images in the full database is calculated. To achieve a correct identification, the similarity of the model's representation has to be highest with one of the images showing the correct person. 

\paragraph{Single Reference Evaluation (Single-Ref).} The second scenario is more challenging. In this case, the attacker has only a single image per person. Hence, the model has to retrieve the single correct image from the database. 

\paragraph{Few Reference Evaluation (Few-Ref).} In the last scenario, the database consists of a at most a few images per person, which is less challenging than the single reference evaluation but more challenging than the full reference evaluation. We set the maximum number of images per person to 5 for our few reference evaluation.

The three described protocols evaluate the results for different attacker strengths. The more images per person the attacker has, the easier it is to identify the persons.
We report top 1 and top $k$ scores for multiple values $k$ for all three protocols.

%% file: sec/4_results.tex
\section{Results}

\subsection{Dataset}

We use the publicly available Large-scale CelebFaces Attributes (CelebA) dataset \cite{celeba} for our experiments. It consists of 202,599 face images from 10,177 individuals. The number of images per individual ranges from 1 to 35. 

\begin{figure}[htb]
  \centering
  \begin{subfigure}{0.32\linewidth}
  \centering
    \includegraphics[width=\linewidth]{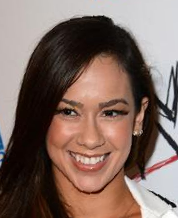}
    \end{subfigure}
    \begin{subfigure}{0.32\linewidth}
    \centering
    \includegraphics[width=\linewidth]{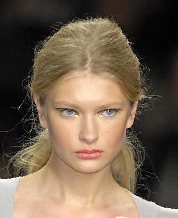}
    \end{subfigure}
    \begin{subfigure}{0.32\linewidth}
  \centering
    \includegraphics[width=\linewidth]{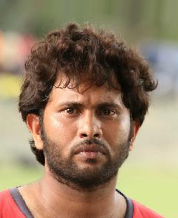}
    \end{subfigure}
    \par\vspace{0.1cm}
    \begin{subfigure}{0.32\linewidth}
    \centering
    \includegraphics[width=\linewidth]{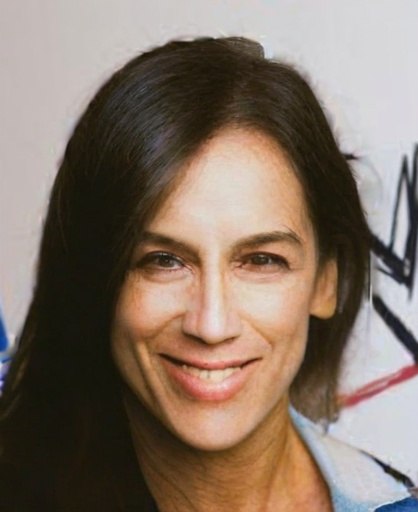}
    \end{subfigure}
    \begin{subfigure}{0.32\linewidth}
  \centering
    \includegraphics[width=\linewidth]{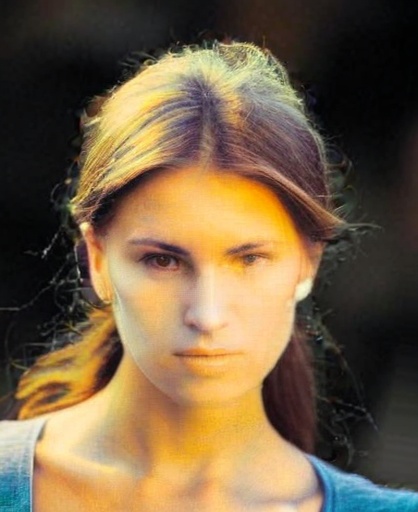}
    \end{subfigure}
    \begin{subfigure}{0.32\linewidth}
    \centering
    \includegraphics[width=\linewidth]{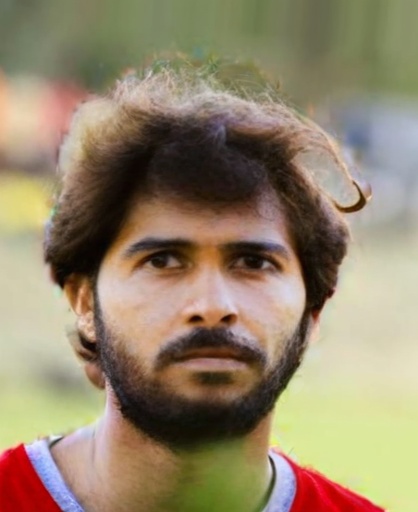}
    \end{subfigure}
   \caption{Qualitative results for original images (top row) and augmented images (bottom row) from our dataset which is based on CelebA \cite{celeba}.}
   \label{fig:instanceaug}
\end{figure}

\paragraph{Augmentations.} We apply Instance Augmentation \cite{instance_aug} to most images of CelebA. We detect the person's bounding box using GroundingDINO \cite{groundingdino} and segmentation masks using SAM \cite{sam} as suggested by the authors of Instance Augmentation. However, some masks were not correct and some created augmentations did not pass the NSFW filter included in the Instance Augmentation pipeline. After removing all such images, our dataset consists of 189,582 images. We provide some qualitative examples for images and their augmentations in \cref{fig:instanceaug}.
\begin{table}
  \centering
  \begin{tabular}{lccc}
    \toprule
    Split & Persons & Original & Augmented \\
    \midrule
    Train & 3654 & 100217 & 154672\\
    Validation & 1466 & 20697 & 10368\\
    Test & 3421 & 48226 & 24542\\
    \bottomrule
  \end{tabular}
  \caption{Statistics of our dataset. The table shows the number of individuals, the number of original images, and the number of augmented images for each subset.}
  \label{tab:dataset}
\end{table}

\begin{table*}[htb]
  \centering
  \begin{tabular}{c|ccc|ccc|ccc}
    \toprule
    & \multicolumn{3}{c|}{Full-Ref} & \multicolumn{3}{c|}{Few-Ref} & \multicolumn{3}{c}{Single-Ref} \\
    BS & top 1 & top 5 & top 10 & top 1 & top 5 & top 10 & top 1 & top 10 & top 100  \\
    \midrule
    16  & 39.0 & 58.4 & 66.2 & 41.4 & 60.5 & 68.1 & 17.7 & 37.0 & 67.5\\
    32  & 56.9 & 73.9 & 79.6 & 56.4 & 72.3 & 78.1 & 22.7 & 42.3 & 70.0\\
    64  & 58.5 & 73.5 & 78.8 & 57.2 & 71.7 & 76.9 & 21.9 & 40.7 & 68.0 \\
    128 & 68.1 & 81.2 & 85.3 & 64.5 & 76.9 & 81.1 & 23.9 & 41.4 & 68.5 \\
    \bottomrule
  \end{tabular}
  \caption{Evaluation scores in \% for different batch sizes for all three evaluation protocols mentioned in \cref{sec:eval}. Depending on the protocol, we evaluate different top $k$ scores.}
  \label{tab:results}
\end{table*}

\paragraph{Data Splits}\label{sec:datasplits}
We split the dataset into training, validation, and test set. Unless otherwise specified, we ensure that all images from the same person (anonymized or not) are contained in the same split. This ensures that the model does not benefit if it overfits on a person from the training set. Note that even in Full-Ref evaluation, the evaluation is conducted on images from persons that the model did not see at all during training. The different evaluation scenarios only cover the amount of original images that are available per unknown person.
Individuals with many available images (augmented and original) are preferably moved to the training set, as more available images from the same individual are beneficial for training. See \cref{tab:dataset} for our dataset statistics.

\subsection{Contrastive Learning}

We set the temperature constant $\tau = 0.05$ and use an AdamW optimizer with a learning rate of $10^{-4}$ during training unless mentioned otherwise. Since we want to show that simple models are sufficient to achieve good results in this task, we use a ResNet50 \cite{resnet} backbone as in the original SimCLR method \cite{simclr}. We select the best weights according to our validation set and report results on the test set, which consists of 3,421 unique persons with corresponding 24,542 augmented and 48,226 original images (see \cref{tab:dataset}). Hence, in the Full-Ref evaluation protocol, the attackers database consists of all 48,226 original images. In the Single-Ref evaluation scenario, the database contains only 3,421 images, which is only approx. 7\% of the full database size. For Few-Ref evaluation, the database size is 15,880 images, which is approx. 33\% of the full DB size. In Few-Ref, at most 5 images per person are used, for some persons fewer than 5 images are available.

We present the results for our model trained with different batch sizes in \cref{tab:results}. Batch size 128 achieves the best results for most evaluations. It identifies 68.1\% of the persons correctly provided with the full database and for 81.2\% of the examples, the correct person is in its top three outputs. These results are remarkable, since the total amount of persons is over 3k. Limiting the database to a single image per person, the results are worse for all models. However, our best model still identifies 23.9\% correctly on first try and 42.3\% in the top 10. 

Larger batch sizes have a large influence on evaluations with more images. Single-Ref scores vary slightly, Full-Ref scores significantly. Few-Ref evaluation results are pretty close to Full-Ref results, especially for batch size 32. The Single-Ref results have a large gap to the other evaluation scenarios. Hence, an attacker profits from collecting a few images per person, but collecting large amounts of images per person is not necessary to achieve good results.

Although batch size 128 shows the best results for most experiments, we are unable to test even larger batch sizes since they consume large amounts of computational resources. For the same reason, we select batch size 32 for further ablation studies, since it provides a good trade-off between performance and required computational resources. It surpasses batch size 64 in most of our experiments and even batch size 128 in top 10 and top 100 scores for Single-Ref evaluation.

\subsection{Ablation Studies}

We assume that the edges of the persons in the augmented images are important that our method works. To further investigate this assumption, we conduct different ablation studies regarding the importance of edges. Moreover, we experiment with different backbones and different batch sizes.

\subsubsection{Influence of Conditioning during Augmentation}\label{sec:ablation_conditioning}

We want to investigate the importance of edges for our approach. Therefore, we compare the results of our model trained with differently augmented images:
\begin{enumerate}
    \item Standard Instance Augmentation using depth and edges information
    \item Instance Augmentation using only depth information, with \textbf{no} edge information
    \item Instance Augmentation using only edges information, with \textbf{no} depth information
    \item Instance Augmentation using only segmentation masks
\end{enumerate}
Since Instance Augmentation is computationally expensive, we evaluate these experiments on smaller datasets with approx. 20k images. Each dataset consists of approximately the same amounts of augmented and original images and individuals. 

\paragraph{Results.} 
The results are shown in \cref{tab:ablation}. These scores are not directly comparable to our base model, as we use significantly smaller datasets for training and evaluation. The evaluation dataset consists of approximately 100 individuals and around 3,000 images. We observe that conditioning the augmentations only on depth or only on edges results in slightly lower performance compared to the standard setting, which includes both conditions. However, the decrease in performance is not as significant as anticipated.
\begin{table}[htb]
  \centering
  \resizebox{\linewidth}{!}{ 
  \begin{tabular}{cccrrrr}
    \toprule
    \multirow{2}{*}{Depth} & \multirow{2}{*}{Edges} & \multirow{2}{*}{Segm.} &\multicolumn{2}{c}{Full-Ref} & \multicolumn{2}{c}{Single-Ref} \\
    \cmidrule(lr){4-5}\cmidrule(lr){6-7}
    & & & top 1 & top 10 & top 1 & top 10 \\
    \midrule
    \checkmark & \checkmark & \checkmark & 72.1 & 91.0 & 22.4 & 60.0 \\
    \checkmark & - & \checkmark & 65.8 & 87.7 & 18.1 & 56.0\\
    - & \checkmark & \checkmark & 64.0 & 86.6 & 17.9 & 53.6 \\
    - & - & \checkmark & 21.7 & 44.3 & 3.2 & 14.9 \\
    \bottomrule
  \end{tabular}
  }
  \caption{Ablation study on Instance Augmentation conditioning. We provide experiments for augmented images conditioned on depth and edges, depth only, edges only, and none of both. Conditioning on the segmentation mask (Segm.) is always included. Results include top-1 and top-10 accuracy in \% for Full-Ref and Single-Ref evaluation.}
  \label{tab:ablation}
\end{table}
\begin{figure*}[ht]
  \centering
  \begin{subfigure}{0.19\linewidth}
    \includegraphics[width=0.9\linewidth]{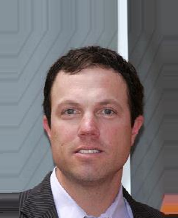}
    \end{subfigure}
    \begin{subfigure}{0.19\linewidth}
    \includegraphics[width=0.9\linewidth]{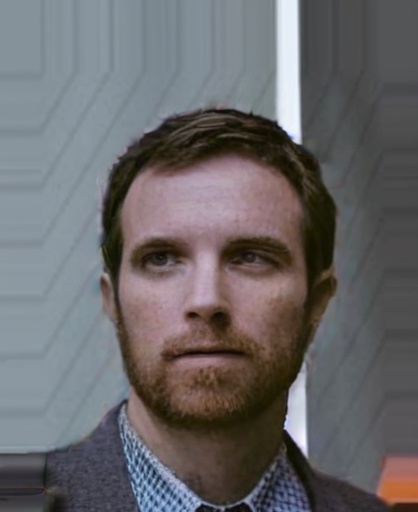}
    \end{subfigure}
    \begin{subfigure}{0.19\linewidth}
    \includegraphics[width=0.9\linewidth]{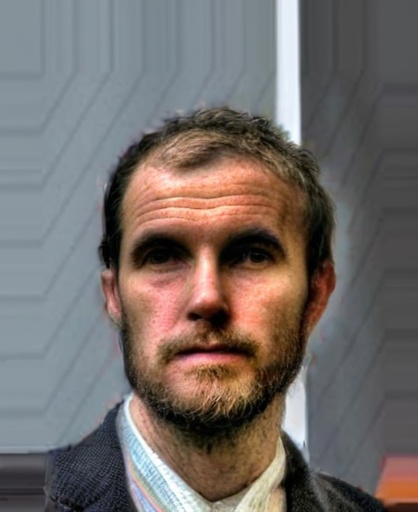}
    \end{subfigure}
    \begin{subfigure}{0.19\linewidth}
    \includegraphics[width=0.9\linewidth]{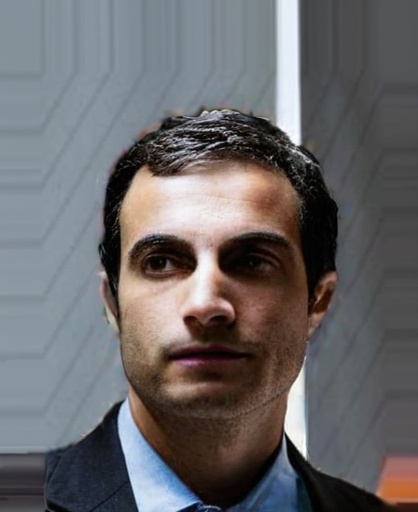}
    \end{subfigure}
    \begin{subfigure}{0.19\linewidth}
    \includegraphics[width=0.9\linewidth]{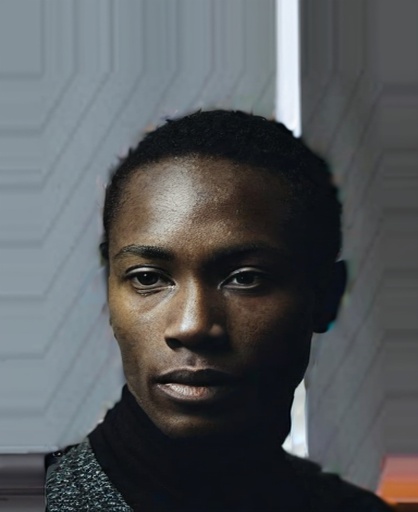}
    \end{subfigure}
    \par\vspace{0.2cm}
    \begin{subfigure}{0.19\linewidth}
    \includegraphics[width=0.9\linewidth]{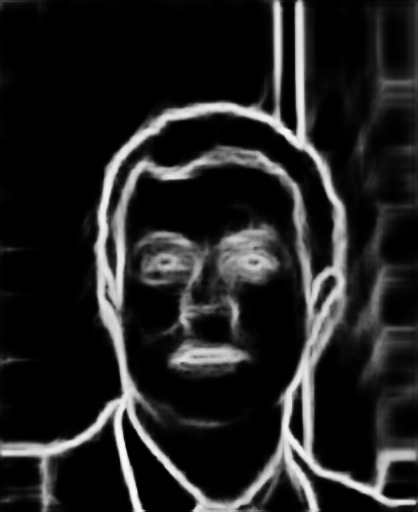}
    \caption{original}
    \end{subfigure}
    \begin{subfigure}{0.19\linewidth}
    \includegraphics[width=0.9\linewidth]{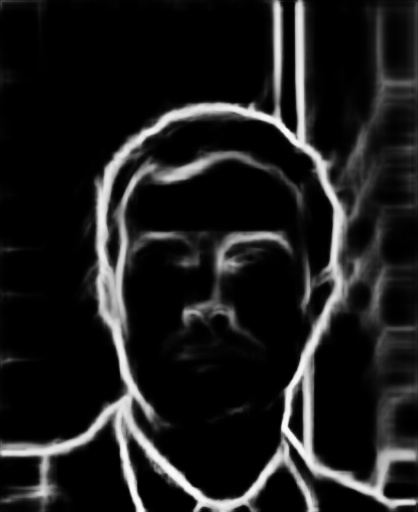}
    \caption{depth and edges}
    \end{subfigure}
    \begin{subfigure}{0.19\linewidth}
    \includegraphics[width=0.9\linewidth]{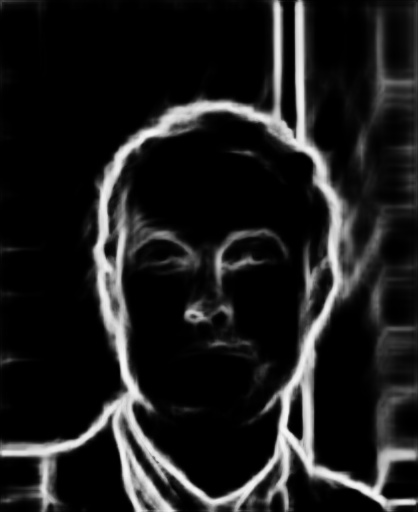}
    \caption{depth}
    \end{subfigure}
    \begin{subfigure}{0.19\linewidth}
    \includegraphics[width=0.9\linewidth]{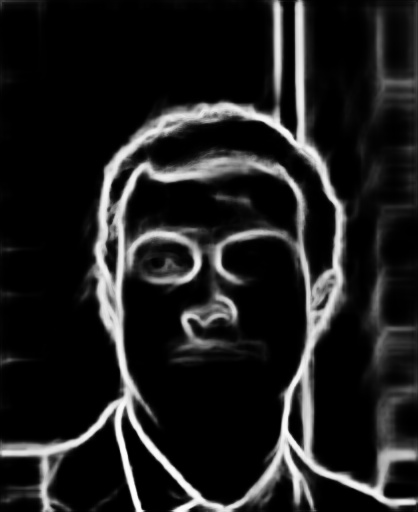}
    \caption{edges}
    \end{subfigure}
    \begin{subfigure}{0.19\linewidth}
    \includegraphics[width=0.9\linewidth]{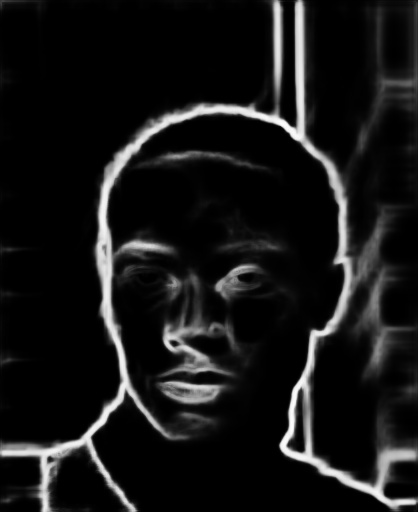}
    \caption{segmentation only}
    \end{subfigure}
   \caption{Qualitative examples for augmented images with different conditioning based on CelebA \cite{celeba} (Top) and corresponding edges detected with a HED detector \cite{hed} (Bottom).}
   \label{fig:conditioning_aug}
\end{figure*}
Upon visually inspecting the augmented images, we find that depth-only conditioned images preserve the edges almost as well as those conditioned on edges alone or both conditions. Based on this, we also conduct an experiment with images conditioned solely on the segmentation mask. Examples of augmented images for all conditioning variants are shown in \cref{fig:conditioning_aug}. Conditioning only on the segmentation mask leads to the worst results by a large margin, with only 3.2\% of individuals correctly identified in the Single-Ref evaluation.

For further edge analysis, we compute the edges of all datasets using a HED detector \cite{hed}. Examples of these images are provided in the bottom row of \cref{fig:conditioning_aug}. We then compare the Structural Similarity Index (SSIM) and the mean L1 distance to the edges of the original images. The results, shown in \cref{tab:edge_similarity}, reveal that the standard augmentation with both conditioning factors is the most similar to the original edges, though the difference is minimal compared to conditioning on edges-only or depth-only. This supports our visual observation that depth and edge conditioning preserve edges similarly well. Omitting both conditionings results in images with significantly lower edge similarity.
\begin{table}[htb]
  \centering
  \begin{tabular}{ccccc}
    \toprule
    Depth & Edges & Segm. & SSIM $\uparrow$ & L1 $\downarrow$\\
    \midrule
    \checkmark & \checkmark & \checkmark & 0.62 & 0.086 \\
    \checkmark & - & \checkmark & 0.59 & 0.092 \\
    - & \checkmark & \checkmark & 0.61 & 0.094 \\
    - & - & \checkmark & 0.51 & 0.128 \\
    \bottomrule
  \end{tabular}
  \caption{SSIM and mean L1 scores for edge similarity of differently augmented images compared to the original images.}
  \label{tab:edge_similarity}
\end{table}

\subsubsection{Architecture}

Apart from studying the influence of augmentation conditioning, we also conduct ablation studies regarding the temperature parameter of our contrastive learning pipeline. Furthermore, we investigate the usage of other backbone architectures.

\paragraph{Temperature.} 

\begin{table}[htb]
  \centering
  \begin{tabular}{rrrrr}
    \toprule
    & \multicolumn{2}{c}{Full-Ref} & \multicolumn{2}{c}{Single-Ref} \\
    \cmidrule(lr){2-3}\cmidrule(lr){4-5}
    $\tau$ & top 1 & top 10 & top 1 & top 10 \\
    \midrule
    1    &  0.7 &  5.2 &  0.8 &  5.6\\
    0.5  &  3.3 & 14.9 &  2.8 & 13.4\\
    0.1  & 33.9 & 60.5 & 16.3 & 36.1\\
    0.05 & 56.9 & 79.6 & 22.7 & 42.3\\
    0.01 & 61.1 & 80.7 & 22.1 & 39.3\\
    \bottomrule
  \end{tabular}
  \caption{Evaluation scores in \% for different temperature values.}
  \label{tab:temp}
\end{table}

The temperature parameter ($\tau$) plays a critical role in the similarity calculation. Lower values of $\tau$ increase the separation between similarities, making it harder for the model to pull matching pairs closer together. The results, shown in \cref{tab:temp}, demonstrate that using $\tau = 1$ yields extremely poor performance, highlighting the importance of this parameter during training. As the temperature decreases, the training duration until convergence increases. Interestingly, our model performs best with lower temperature values, which contrasts with the findings of SimCLR \cite{simclr}. In their setting, $\tau = 0.1$ achieved the best results, and further decreasing $\tau$ led to worse performance. In our case, the model trained with $\tau = 0.05$ delivers the best scores for the Single-Ref evaluation, while the model trained with $\tau = 0.01$ achieves the best results for the Full-Ref evaluation.

\paragraph{Backbone.}
Our main goal is to show that a simple model is capable of identifying persons in instance augmented images of the persons' faces. Therefore, we used ResNet50 as a backbone so far, since it is a simple, easy to use backbone. However, we are interested if we can tune the performance of our method further when using a more recent backbone. Hence, we choose ConvNeXt \cite{convnext} and ViT \cite{vit} pretrained with DINOv2 \cite{dinov2} as alternative backbones. We select model variants with a more or less similar amount of parameters compared to ResNet50, resulting in ConvNeXt tiny and ViT-S/14.
We execute experiments with batch size 32 and 128 for both variants and show the results in \cref{tab:backbones}. The number of parameters per model is also included in the table.
\begin{table}[htb]
  \centering
  \resizebox{\linewidth}{!}{ 
  \begin{tabular}{rrrrrrr}
    \toprule
    \multirow{2}{*}{Backbone} & \multirow{2}{*}{Param} & &\multicolumn{2}{c}{Full-Ref} & \multicolumn{2}{c}{Single-Ref} \\
    \cmidrule(lr){4-5}\cmidrule(lr){6-7}
    & & BS & top 1 & top 10 & top 1 & top 10 \\
    \midrule
    \multirow{2}{*}{ResNet}   & \multirow{2}{*}{26\,M}   &  32 & 56.9 & 79.6 & 22.7 & 42.3\\
                            &     & 128 & 68.1 & 85.3 & 23.9 & 41.4\\
                                 \cmidrule(lr){1-7}
    \multirow{2}{*}{ConvNeXt}& \multirow{2}{*}{29\,M}    &  32 & 55.2 & 78.4 & 23.1 & 43.6 \\
                             &    & 128 & 69.7 & 88.6 & 26.3 & 46.1\\
                                 \cmidrule(lr){1-7}
    \multirow{2}{*}{ViT-S} & \multirow{2}{*}{21\,M}      & 32 & 46.6 & 72.4 & 21.2 & 42.0\\
                             &    & 128 & 53.2 & 77.1 & 23.5 & 44.5\\
                             \cmidrule(lr){1-7}
    \multirow{2}{*}{ViT-B} & \multirow{2}{*}{86\,M}      & 32 & 58.6 & 80.7 & 23.9 & 44.9\\
                              &   & 64 & 61.4 & 82.3 & 25.3 & 45.7\\
    \bottomrule
  \end{tabular}
  }
  \caption{Evaluation scores in \% for different backbone architectures: ResNet50, ConvNeXt-tiny, ViT-S/14, and ViT-B/14.}
  \label{tab:backbones}
\end{table}

As expected, modern backbones further improve performance. ConvNeXt with a batch size of 128 outperforms the ResNet50 model by 1.6\% in top-1 accuracy for Full-Ref evaluation and by 1.4\% for Single-Ref evaluation. ConvNeXt appears to benefit more from a larger batch size, showing a greater improvement over its batch size 32 variant compared to ResNet. For Full-Ref evaluation, ResNet with a batch size of 32 outperforms ConvNeXt.
Despite being pretrained with DINOv2, the ViT model performs worse than the convolutional models on most evaluation metrics. For a batch size of 128, ViT outperforms ResNet in top-10 accuracy, though its top-1 score is slightly lower. One possible reason is that ViT-S has 21M parameters, 19\% fewer than ResNet50. To further explore the potential of a ViT backbone, we train a ViT-B model with 86M parameters, over three times the size of ResNet50. While this comparison isn’t entirely fair, it provides insight into the ViT’s capabilities. As shown in \cref{tab:backbones}, the larger ViT-B model yields better results. For batch size 32, it outperforms all other models trained with the same batch size. Due to its larger size, we cannot train ViT-B with batch size 128 but include a training with batch size 64 to assess the impact of batch size on performance. The results show that a larger batch size improves the outcome, even for larger models. However, using an even larger batch size with smaller models results in better performance.

\paragraph{Data Split.} In this experiment, we investigate the attacker's strength in a scenario where the attacker's model can be trained using augmented images of the target persons. 
Hence, we distribute images from the same person across the training, validation, and test set.
An attacker who has access to the anonymization pipeline might use this to their advantage and fine tune a model on a certain person.
This experiment emulates this scenario by allowing models to remember people from the training set.

%% file: sec/5_blackbox.tex
\section{Black-Box Attack: Identifying Individuals Using Edge Images Only}\label{sec:blackbox}

The previously discussed methods assume that an attacker has access to anonymized images.
We extend our analysis to a black box attack where an attacker does not require any anonymized images or access to the anonymization pipeline.

\begin{figure}[htb]
    \centering
    \begin{subfigure}{0.3\linewidth}
        \centering
        \includegraphics[width=\linewidth]{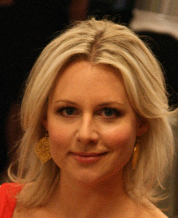}
    \end{subfigure}
    \hfill
    \begin{subfigure}{0.3\linewidth}
        \centering
        \includegraphics[width=\linewidth]{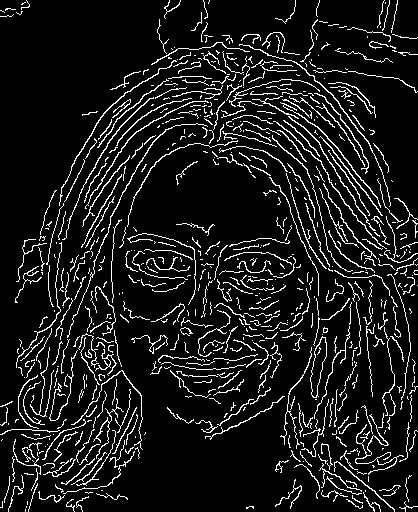}
    \end{subfigure}
    \hfill
    \begin{subfigure}{0.3\linewidth}
        \centering
        \includegraphics[width=\linewidth]{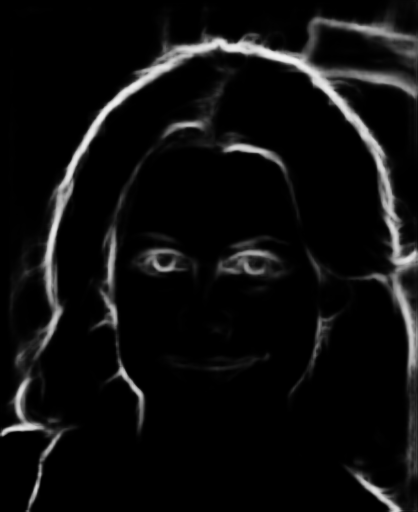}
    \end{subfigure}
   \caption{Edge transformations of the original image (left) using Canny edge detector \cite{canny} (middle) and HED \cite{hed} (right).}
   \label{fig:blackbox_edges}
\end{figure}

The analyzed anonymization strategy uses latent diffusion models that are conditioned on edges and depth from the original image.
Therefore, it is instructed to preserve these edges in the anonymized version.
As we already investigated in \cref{sec:ablation_conditioning}, the preservation of the edges is the main reason why our model is successful. In this section, we make further use of this fact and train a contrastive learning model on non-anonymized images that are transformed to edge images.
To retrieve a person during testing, we transform the anonymized image to its edge representation and then apply the trained model.
Hence, the model sees no anonymized image during training, but is applied to anonymized images during evaluation.
As before, we make sure that all images from the same person are either in the training set or in the test set but not in both.
We run experiments using Canny edge detection \cite{canny} combined with a $5 \times 5$ Gaussian blur, and using HED \cite{hed} to transform the original images to edge images.
See \cref{fig:blackbox_edges} for example edge images.

\begin{table}[htb]
  \centering
  \begin{tabular}{rrrrrr}
    \toprule
    \multirow{2}{*}{Edges} & \multirow{2}{*}{BB} &\multicolumn{2}{c}{Full-Ref} & \multicolumn{2}{c}{Single-Ref} \\
    \cmidrule(lr){3-4}\cmidrule(lr){5-6}
    & & top 10 & top 100 & top 10 & top 100 \\
    \midrule
    Canny & \checkmark & 0.6 & 4.3 & 0.6 & 5.8\\
    HED & \checkmark & 4.9 & 18.7 & 4.6 & 20.6\\
    \midrule
    Canny & -&  24.2 & 54.9 & 16.9 & 42.5\\ %
    HED &-&  50.5 & 76.6  & 27.2 & 54.1 \\ %
    \bottomrule
  \end{tabular}
  \caption{Evaluation scores in \% for our training on edge images. The top half displays the results for the black box (BB) attack, the bottom half the results for training on the edges of augmented images.}
  \label{tab:blackbox}
\end{table}

The results are presented in the top half of \cref{tab:blackbox}.  
Even without any anonymized images during training, the HED-based approach correctly identifies the target individual in the top-10 results with a probability of 4.9\%. This is remarkable compared to random guessing, which would yield an accuracy of approximately 0.03\%. The Canny-based approach performs significantly worse but remains 20 times more effective than random guessing.  

To further investigate the importance of edge preservation, we conduct a similar experiment but allow access to anonymized images during training. In this setting, we train on edge images extracted from augmented images. The results, shown in the bottom half of \cref{tab:blackbox}, indicate that access to augmented images or the augmentation pipeline drastically improves the attacker's success. HED-based edges again yield superior performance, revealing the identity of 50.5\% of individuals in the top-10 Full-Ref evaluation. While the absolute score remains lower than training directly on augmented images, these findings demonstrate that edges preserve critical identity-related information, making them a key factor in the success of our method.

%% file: sec/6_conclusion.tex
\section{Conclusion}

In this work we analyzed recent work on using conditioned latent diffusion models as an anonymization strategy.
In contrast to prior claims, we demonstrate that a simple contrastive learning approach is sufficient to reliably retrieve the anonymized person from a pool of candidates.

We analyze the performance of contrastive learning for person retrieval finding that a model trained on original and anonymized image pairs is capable to retrieve the correct person out of \todo{??} candidates with a probability of \todo{68\%}.
We provide a thorough analysis of the trained model and show that the performance depends on the fact that many edges are preserved during the anonymization process.
Although the person might seem different to the human eye, a neural network can use the features of the edges to correctly disambiguate between the different people.

Furthermore, we show that preserved image features like edges enable a black box attack where the attacker does not require any access to anonymized images or the anonymization pipeline.
We train a model directly on edge images and show that this is sufficient to retrieve people from a set of candidates.

Using conditioned latent diffusion models is a great way to increase training data and improve model performance.
However, based on our findings, we advice against using methods that preserve features like edges for data anonymization.

%% file: main.bbl
\begin{thebibliography}{38}
\providecommand{\natexlab}[1]{#1}
\providecommand{\url}[1]{\texttt{#1}}
\expandafter\ifx\csname urlstyle\endcsname\relax
  \providecommand{\doi}[1]{doi: #1}\else
  \providecommand{\doi}{doi: \begingroup \urlstyle{rm}\Url}\fi

\bibitem[Canny(1986)]{canny}
John Canny.
\newblock A computational approach to edge detection.
\newblock \emph{IEEE Transactions on Pattern Analysis and Machine Intelligence}, PAMI-8\penalty0 (6):\penalty0 679--698, 1986.

\bibitem[Cao et~al.(2022)Cao, Liu, Wen, Zhu, Xie, Song, Li, and Yin]{9828699}
Jingyi Cao, Bo Liu, Yunqian Wen, Yunhui Zhu, Rong Xie, Li Song, Lin Li, and Yaoyao Yin.
\newblock Hiding among your neighbors: Face image privacy protection with differential private k-anonymity.
\newblock In \emph{2022 IEEE International Symposium on Broadband Multimedia Systems and Broadcasting (BMSB)}, pages 1--6, 2022.

\bibitem[Cao et~al.(2024)Cao, Chen, Liu, Ding, Xie, Song, Li, and Zhang]{face_deidentify}
Jingyi Cao, Xiangyi Chen, Bo Liu, Ming Ding, Rong Xie, Li Song, Zhu Li, and Wenjun Zhang.
\newblock Face de-identification: State-of-the-art methods and comparative studies, 2024.

\bibitem[Caron et~al.(2020)Caron, Misra, Mairal, Goyal, Bojanowski, and Joulin]{NEURIPS2020_70feb62b}
Mathilde Caron, Ishan Misra, Julien Mairal, Priya Goyal, Piotr Bojanowski, and Armand Joulin.
\newblock Unsupervised learning of visual features by contrasting cluster assignments.
\newblock In \emph{Advances in Neural Information Processing Systems}, pages 9912--9924. Curran Associates, Inc., 2020.

\bibitem[Chen et~al.(2020{\natexlab{a}})Chen, Kornblith, Norouzi, and Hinton]{chen2020simpleframeworkcontrastivelearning}
Ting Chen, Simon Kornblith, Mohammad Norouzi, and Geoffrey Hinton.
\newblock A simple framework for contrastive learning of visual representations, 2020{\natexlab{a}}.

\bibitem[Chen et~al.(2020{\natexlab{b}})Chen, Kornblith, Norouzi, and Hinton]{simclr}
Ting Chen, Simon Kornblith, Mohammad Norouzi, and Geoffrey Hinton.
\newblock A simple framework for contrastive learning of visual representations.
\newblock In \emph{International conference on machine learning}, pages 1597--1607. PmLR, 2020{\natexlab{b}}.

\bibitem[Deng et~al.(2019)Deng, Guo, Xue, and Zafeiriou]{arcface}
Jiankang Deng, Jia Guo, Niannan Xue, and Stefanos Zafeiriou.
\newblock Arcface: Additive angular margin loss for deep face recognition.
\newblock In \emph{2019 IEEE/CVF Conference on Computer Vision and Pattern Recognition (CVPR)}, pages 4685--4694, 2019.

\bibitem[Dosovitskiy et~al.(2020)Dosovitskiy, Beyer, Kolesnikov, Weissenborn, Zhai, Unterthiner, Dehghani, Minderer, Heigold, Gelly, et~al.]{vit}
Alexey Dosovitskiy, Lucas Beyer, Alexander Kolesnikov, Dirk Weissenborn, Xiaohua Zhai, Thomas Unterthiner, Mostafa Dehghani, Matthias Minderer, Georg Heigold, Sylvain Gelly, et~al.
\newblock An image is worth 16x16 words: Transformers for image recognition at scale.
\newblock \emph{arXiv preprint arXiv:2010.11929}, 2020.

\bibitem[Gadotti et~al.(2024)Gadotti, Rocher, Houssiau, Creţu, and de~Montjoye]{anonymization_survey}
Andrea Gadotti, Luc Rocher, Florimond Houssiau, Ana-Maria Creţu, and Yves-Alexandre de Montjoye.
\newblock Anonymization: The imperfect science of using data while preserving privacy.
\newblock \emph{Science Advances}, 10\penalty0 (29):\penalty0 eadn7053, 2024.

\bibitem[Guo et~al.(2016)Guo, Zhang, Hu, He, and Gao]{msceleb}
Yandong Guo, Lei Zhang, Yuxiao Hu, X. He, and Jianfeng Gao.
\newblock Ms-celeb-1m: A dataset and benchmark for large-scale face recognition.
\newblock In \emph{ECCV}, 2016.

\bibitem[Hanisch et~al.(2024)Hanisch, Todt, Patino, Evans, and Strufe]{face_attack}
Simon Hanisch, Julian Todt, Jose Patino, Nicholas Evans, and Thorsten Strufe.
\newblock A false sense of privacy: Towards a reliable evaluation methodology for the anonymization of biometric data.
\newblock \emph{Proceedings on Privacy Enhancing Technologies}, 2024.

\bibitem[Hasan et~al.(2018)Hasan, Hassan, Li, Caine, Crandall, Hoyle, and Kapadia]{viewer_experience}
Rakibul Hasan, Eman Hassan, Yifang Li, Kelly Caine, David~J. Crandall, Roberto Hoyle, and Apu Kapadia.
\newblock Viewer experience of obscuring scene elements in photos to enhance privacy.
\newblock In \emph{Proceedings of the 2018 CHI Conference on Human Factors in Computing Systems}, page 1–13, New York, NY, USA, 2018. Association for Computing Machinery.

\bibitem[He et~al.(2016)He, Zhang, Ren, and Sun]{resnet}
Kaiming He, Xiangyu Zhang, Shaoqing Ren, and Jian Sun.
\newblock Deep residual learning for image recognition.
\newblock In \emph{Proceedings of the IEEE conference on computer vision and pattern recognition}, pages 770--778, 2016.

\bibitem[He et~al.(2020)He, Fan, Wu, Xie, and Girshick]{He_2020_CVPR}
Kaiming He, Haoqi Fan, Yuxin Wu, Saining Xie, and Ross Girshick.
\newblock Momentum contrast for unsupervised visual representation learning.
\newblock In \emph{Proceedings of the IEEE/CVF Conference on Computer Vision and Pattern Recognition (CVPR)}, 2020.

\bibitem[Hill et~al.(2016)Hill, Zhou, Saul, and Shacham]{privacy_blurring}
Steven Hill, Zhimin Zhou, Lawrence Saul, and Hovav Shacham.
\newblock On the (in)effectiveness of mosaicing and blurring as tools for document redaction.
\newblock \emph{Proceedings on Privacy Enhancing Technologies}, 2016, 2016.

\bibitem[Hukkelås et~al.(2019)Hukkelås, Mester, and Lindseth]{hukkelås2019deepprivacygenerativeadversarialnetwork}
Håkon Hukkelås, Rudolf Mester, and Frank Lindseth.
\newblock Deepprivacy: A generative adversarial network for face anonymization, 2019.

\bibitem[Jaiswal et~al.(2021)Jaiswal, Babu, Zadeh, Banerjee, and Makedon]{technologies9010002}
Ashish Jaiswal, Ashwin~Ramesh Babu, Mohammad~Zaki Zadeh, Debapriya Banerjee, and Fillia Makedon.
\newblock A survey on contrastive self-supervised learning.
\newblock \emph{Technologies}, 9\penalty0 (1), 2021.

\bibitem[Kirillov et~al.(2023)Kirillov, Mintun, Ravi, Mao, Rolland, Gustafson, Xiao, Whitehead, Berg, Lo, et~al.]{sam}
Alexander Kirillov, Eric Mintun, Nikhila Ravi, Hanzi Mao, Chloe Rolland, Laura Gustafson, Tete Xiao, Spencer Whitehead, Alexander~C Berg, Wan-Yen Lo, et~al.
\newblock Segment anything.
\newblock In \emph{Proceedings of the IEEE/CVF international conference on computer vision}, pages 4015--4026, 2023.

\bibitem[Kupyn and Rupprecht(2024)]{instance_aug}
Orest Kupyn and Christian Rupprecht.
\newblock Dataset enhancement with instance-level augmentations.
\newblock \emph{arXiv preprint arXiv:2406.08249}, 2024.

\bibitem[Li and Choi(2021)]{privacy_imggen}
Tao Li and Min~Soo Choi.
\newblock Deepblur: A simple and effective method for natural image obfuscation, 2021.

\bibitem[Li and Lin(2019)]{privacy_gan}
Tao Li and Lei Lin.
\newblock Anonymousnet: Natural face de-identification with measurable privacy.
\newblock In \emph{2019 IEEE/CVF Conference on Computer Vision and Pattern Recognition Workshops (CVPRW)}, pages 56--65, 2019.

\bibitem[Lin et~al.(2015)Lin, Maire, Belongie, Bourdev, Girshick, Hays, Perona, Ramanan, Zitnick, and Dollár]{coco}
Tsung-Yi Lin, Michael Maire, Serge Belongie, Lubomir Bourdev, Ross Girshick, James Hays, Pietro Perona, Deva Ramanan, C.~Lawrence Zitnick, and Piotr Dollár.
\newblock Microsoft coco: Common objects in context, 2015.

\bibitem[Liu et~al.(2023{\natexlab{a}})Liu, Zeng, Ren, Li, Zhang, Yang, Li, Yang, Su, Zhu, et~al.]{groundingdino}
Shilong Liu, Zhaoyang Zeng, Tianhe Ren, Feng Li, Hao Zhang, Jie Yang, Chunyuan Li, Jianwei Yang, Hang Su, Jun Zhu, et~al.
\newblock Grounding dino: Marrying dino with grounded pre-training for open-set object detection.
\newblock \emph{arXiv preprint arXiv:2303.05499}, 2023{\natexlab{a}}.

\bibitem[Liu et~al.(2023{\natexlab{b}})Liu, Zhang, Hou, Mian, Wang, Zhang, and Tang]{9462394}
Xiao Liu, Fanjin Zhang, Zhenyu Hou, Li Mian, Zhaoyu Wang, Jing Zhang, and Jie Tang.
\newblock Self-supervised learning: Generative or contrastive.
\newblock \emph{IEEE Transactions on Knowledge and Data Engineering}, 35\penalty0 (1):\penalty0 857--876, 2023{\natexlab{b}}.

\bibitem[Liu et~al.(2015)Liu, Luo, Wang, and Tang]{celeba}
Ziwei Liu, Ping Luo, Xiaogang Wang, and Xiaoou Tang.
\newblock Deep learning face attributes in the wild.
\newblock In \emph{Proceedings of International Conference on Computer Vision (ICCV)}, 2015.

\bibitem[Liu et~al.(2022)Liu, Mao, Wu, Feichtenhofer, Darrell, and Xie]{convnext}
Zhuang Liu, Hanzi Mao, Chao-Yuan Wu, Christoph Feichtenhofer, Trevor Darrell, and Saining Xie.
\newblock A convnet for the 2020s.
\newblock \emph{Proceedings of the IEEE/CVF Conference on Computer Vision and Pattern Recognition (CVPR)}, 2022.

\bibitem[McPherson et~al.(2016)McPherson, Shokri, and Shmatikov]{img_attack1}
Richard McPherson, Reza Shokri, and Vitaly Shmatikov.
\newblock Defeating image obfuscation with deep learning, 2016.

\bibitem[Merler et~al.(2019)Merler, Ratha, Feris, and Smith]{merler_faces}
Michele Merler, Nalini Ratha, Rogerio~S. Feris, and John~R. Smith.
\newblock Diversity in faces, 2019.

\bibitem[Mou et~al.(2023)Mou, Wang, Xie, Wu, Zhang, Qi, Shan, and Qie]{t2i_adapter}
Chong Mou, Xintao Wang, Liangbin Xie, Yanze Wu, Jian Zhang, Zhongang Qi, Ying Shan, and Xiaohu Qie.
\newblock T2i-adapter: Learning adapters to dig out more controllable ability for text-to-image diffusion models, 2023.

\bibitem[Oquab et~al.(2023)Oquab, Darcet, Moutakanni, Vo, Szafraniec, Khalidov, Fernandez, Haziza, Massa, El-Nouby, et~al.]{dinov2}
Maxime Oquab, Timoth{\'e}e Darcet, Th{\'e}o Moutakanni, Huy Vo, Marc Szafraniec, Vasil Khalidov, Pierre Fernandez, Daniel Haziza, Francisco Massa, Alaaeldin El-Nouby, et~al.
\newblock Dinov2: Learning robust visual features without supervision.
\newblock \emph{arXiv preprint arXiv:2304.07193}, 2023.

\bibitem[Packhäuser et~al.(2022)Packhäuser, Gündel, Münster, Syben, Christlein, and Maier]{medical_attack}
Kai Packhäuser, Sebastian Gündel, Nicolas Münster, Christopher Syben, Vincent Christlein, and Andreas Maier.
\newblock Deep learning-based patient re-identification is able to exploit the biometric nature of medical chest x-ray data.
\newblock \emph{Scientific Reports}, 12\penalty0 (1):\penalty0 14851, 2022.
\newblock Publisher: Nature Publishing Group.

\bibitem[Ra et~al.(2013)Ra, Govindan, and Ortega]{privacy_p3}
Moo-Ryong Ra, Ramesh Govindan, and Antonio Ortega.
\newblock P3: Toward {Privacy-Preserving} photo sharing.
\newblock In \emph{10th USENIX Symposium on Networked Systems Design and Implementation (NSDI 13)}, pages 515--528, Lombard, IL, 2013. USENIX Association.

\bibitem[Tekli et~al.(2024)Tekli, Al~Bouna, Tekli, Couturier, and Charbel]{img_attack2}
Jimmy Tekli, Bechara Al~Bouna, Gilbert Tekli, Raphaël Couturier, and Antoine Charbel.
\newblock Leveraging deep learning-assisted attacks against image obfuscation via federated learning.
\newblock \emph{Neural Computing and Applications}, 36\penalty0 (25):\penalty0 15667--15684, 2024.

\bibitem[Thakral et~al.(2024)Thakral, Prasad, Aswani, Vatsa, and Singh]{singh_tooner}
Kartik Thakral, Shashikant Prasad, Stuti Aswani, Mayank Vatsa, and Richa Singh.
\newblock Toonergan: Reinforcing gans for obfuscating automated facial indexing.
\newblock In \emph{2024 IEEE/CVF Conference on Computer Vision and Pattern Recognition (CVPR)}, pages 10875--10884, 2024.

\bibitem[Xie and Tu(2015)]{hed}
Saining Xie and Zhuowen Tu.
\newblock Holistically-nested edge detection.
\newblock In \emph{Proceedings of the IEEE international conference on computer vision}, pages 1395--1403, 2015.

\bibitem[Zhai et~al.(2022)Zhai, Guo, Xie, Ma, Wang, and Liu]{privacy_a3gan}
Liming Zhai, Qing Guo, Xiaofei Xie, Lei Ma, Yi~Estelle Wang, and Yang Liu.
\newblock A3gan: Attribute-aware anonymization networks for face de-identification.
\newblock In \emph{Proceedings of the 30th ACM International Conference on Multimedia}, page 5303–5313, New York, NY, USA, 2022. Association for Computing Machinery.

\bibitem[Zhang et~al.(2022)Zhang, Liu, Zhu, Zhou, and Zhou]{attack_survey}
Guangsheng Zhang, Bo Liu, Tianqing Zhu, Andi Zhou, and Wanlei Zhou.
\newblock Visual privacy attacks and defenses in deep learning: a survey.
\newblock \emph{Artificial Intelligence Review}, 55\penalty0 (6):\penalty0 4347--4401, 2022.

\bibitem[Zhang et~al.(2023)Zhang, Rao, and Agrawala]{controlnet}
Lvmin Zhang, Anyi Rao, and Maneesh Agrawala.
\newblock Adding conditional control to text-to-image diffusion models, 2023.

\end{thebibliography}
